\documentclass[letterpaper, 10 pt, conference]{format/ieeeconf}  % Comment this line out if you need a4paper

\IEEEoverridecommandlockouts                              % This command is only needed if 
\usepackage{dsfont}
\usepackage{amsmath}

\usepackage{graphics} % for pdf, bitmapped graphics files
\usepackage{amsmath} % assumes amsmath package installed
\usepackage{amssymb}  % assumes amsmath package installed
\usepackage{todonotes}

\usepackage{amsfonts}
\usepackage{mathtools}
\usepackage[linesnumbered,algoruled,boxed,vlined, noend]{algorithm2e}
\usepackage{amsthm}
\usepackage{comment}
\usepackage{cite}
\usepackage{wrapfig}
\let\labelindent\relax
\usepackage{enumitem}% http://ctan.org/pkg/enumitem
\usepackage{overpic}
\usepackage{xcolor}

\DeclareMathAlphabet{\mathcal}{OMS}{cmsy}{m}{n} % to use old calligraphy with mathptmx

\theoremstyle{definition}

\theoremstyle{remark}

\DeclarePairedDelimiterX{\norm}[1]{\lVert}{\rVert}{#1}

\newcommand{\rom}[1]{\uppercase\expandafter{\romannumeral #1\relax}}

\newcommand{\rect}{\text{$\mathcal{R}$}}

\newcommand{\config}{\text{$\mathcal{X}$}}

\newcommand{\workspace}{\text{$\mathcal{W}$}}
\newcommand{\objects}{\text{$\mathcal{O}$}}
\newcommand{\target}{\text{$\mathcal{T}$}}
\newcommand{\positions}{\text{$\mathcal{P}$}}
\newcommand{\connectedcomp}{\text{$\mathcal{CC}$}}
\newcommand{\pathregion}{\text{$\Pi$}}

{\end{list}}

\usepackage{soul} % only for highlight purposes
\usepackage[a-2b,mathxmp]{pdfx}[2018/12/22]
\usepackage{titlesec}   % Adjust section title spacing

\title{\LARGE \bf
Persistent Homology for Effective Non-Prehensile Manipulation
}

\author{Ewerton R. Vieira, Daniel Nakhimovich, Kai Gao, Rui Wang, Jingjin Yu and Kostas E. Bekris%
\thanks{The authors are with the Department of Computer Science and DIMACS (the Center for Discrete Mathematics and Theoretical Computer Science), Rutgers University, NJ, USA. Email: {\tt\small { \{rw485, kg627, dn332, jy512, kb572\}}@cs.rutgers.edu} and {\tt\small er691@rutgers.edu}.
The work is supported in part by an NSF HDR TRIPODS award 1934924.
}%
}

\begin{document}

\maketitle
\thispagestyle{empty}
\pagestyle{empty}

%%%%%%%%%%%%%%%%%%%%%%%%%%%%%%%%%%%%%%%%%%%%%%%%%%%%%%%%%%%%%%%%%%%%%%%%%%%%%%%%

\begin{abstract} 
This work explores the use of topological tools for achieving effective non-prehensile manipulation in cluttered, constrained workspaces. In particular, it proposes the use of persistent homology as a guiding principle in identifying the appropriate non-prehensile actions, such as pushing, to clean a cluttered space with a robotic arm so as to allow the retrieval of a target object. Persistent homology enables the automatic identification of connected components of blocking objects in the space without the need for manual input or tuning of parameters. The proposed algorithm uses this information to push groups of cylindrical objects together and aims to minimize the number of pushing actions needed to reach to the target. Simulated experiments in a physics engine using a model of the Baxter robot show that the proposed topology-driven solution is achieving significantly higher success rate in solving such constrained problems relatively to state-of-the-art alternatives from the literature. It manages to keep the number of pushing actions low, is computationally efficient and the resulting decisions and motion appear natural for effectively solving such tasks.  
\end{abstract}

\section{Introduction}

%Describe the problem and highlight its importance in robotics
%Explain how humans may be solving it - grouping objects together in some fashion
%Explain how persistent homology may do the same
%State our result and contribution 

In order to retrieve a target object from clutter, a robotic arm may first need to relocate other objects that are blocking access.  Such a task appears often in applications ranging from service robotics (e.g., taking out a can of soda from the fridge) to logistics (e.g., retrieving an ordered product from the shelf of a grocery store). In many cases, especially in homes, the workspace is cluttered and unstructured. But it can still allow a combination of planar, non-prehensile pushes to clear blocking objects and prehensile grasping to pick the target object. This paper focuses on making such strategies for object retrieval more efficient so as to equip robot assistants with this useful skill. 

While there are methods for realizing this robotic skill \cite{dogar2013physics,bejjani2018planning,papallas2020non}, human-level performance in terms of efficiency and smooth, natural motion has yet to be achieved.  When humans perform such tasks, they often perform an implicit ``object grouping''  so as simultaneously push multiple objects and find the least number of pushes before the target can be retrieved. Building on the observation, this work, explores topological tools to systematically identify how objects can be grouped into connected components. Once the objects are grouped the approach identifies pushing actions that are effective in clearing the blocking objects. In particular, persistent homology (PH) provides a persistence diagram for selecting connected components, which group objects that can be pushed together in a single action. 

\begin{figure}[t]
\vspace{1mm}
\centering
\includegraphics[width=0.48\textwidth]{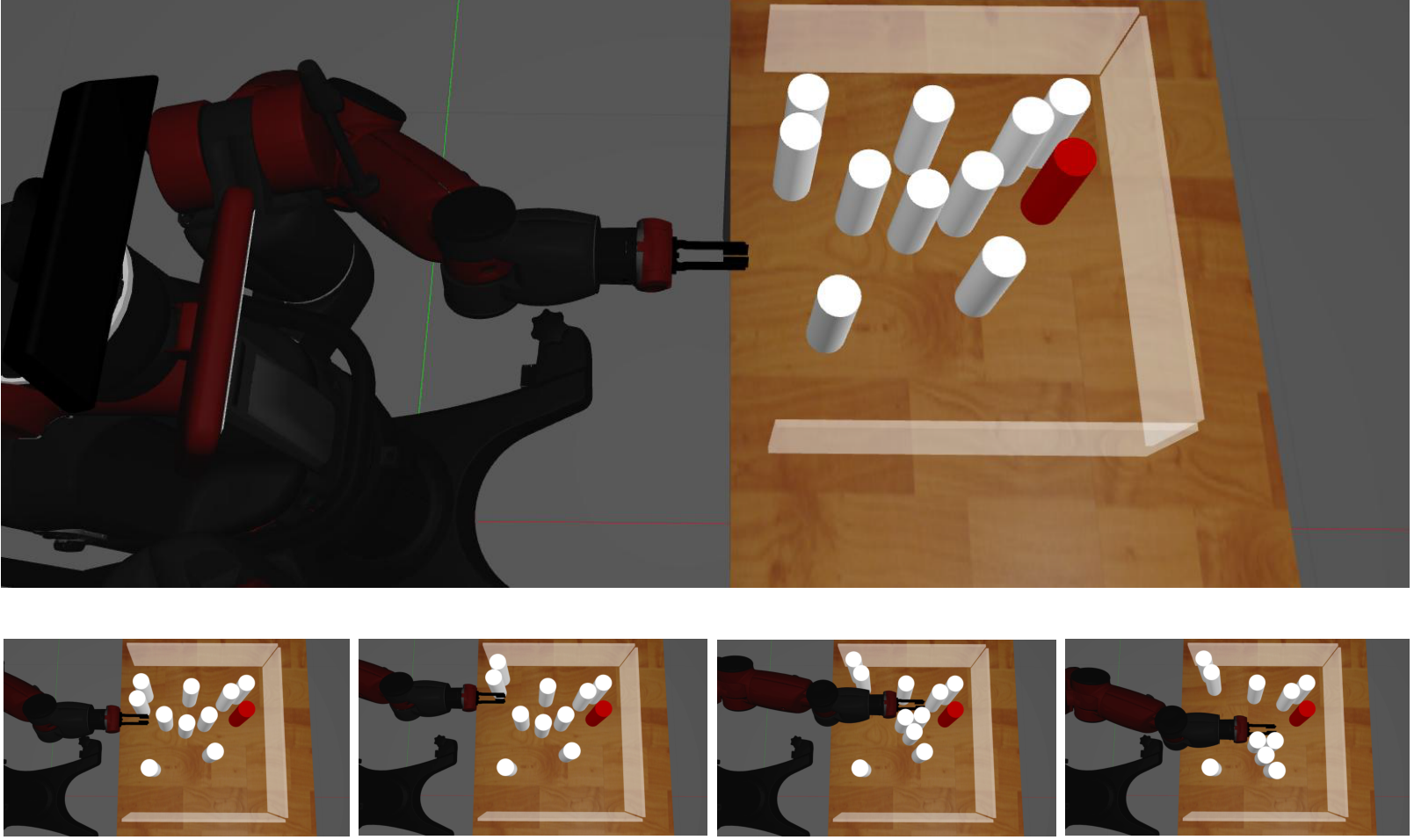}
    \vspace{-.7cm}
    \caption{(top) An example setup where a model of the Baxter robot and a cluttered set of cylindrical objects are modeled in a physics engine. The target object is highlighted and access to it is blocked by the white objects. (bottom) A solution sequence computed by the proposed topology-driven method. The arm rearranges the blocking objects in order to reach its target.}
    \label{fig:setup}
\end{figure}

%It is, thus, not surprising that the manipulation of several objects is often executed by moving one object at a time via grasping.

The dynamics of such pushes are rather complex. A robot arm is usually not designed with push actions in mind. Such operations are in general underactuated, non-prehensile primitives that complicate the achievement of desirable end positions for the objects.  They involve complicated dynamics as objects may collide each other. This work does not aim to understand or model the dynamics of such pushing operations over groups of objects. It shows, however, that reasoning about the topology of the configuration space is still helpful in this context and allows grouping objects that can be relocated effectively under non-prehensile manipulation. In other words, persistent connected components of the objects given by PH can be more reliably pushed together and result in fewer pushes to solve such problems.

Two methods are presented here for selecting connected components of blocking objects in such setups. A. For each configuration instance, the PH Informed Actions approach described in Algorithm \ref{alg:simplest} selects the closest connected component to the arm so that it is the minimal one, which  persists in the persistence diagram (taking into account the restrictions imposed by the size of objects and the arm). B. The PH Informed Search approach described in Algorithm \ref{alg:search} considers all connected components, which persist for different radii in the persistence diagram. It builds a tree with nodes given by the configuration space and it propagates by performing the push action until it gets success or failure. Finally, it searches for the path that leads to success and minimizes the time for performing the pushing actions.

To our knowledge, this is the first application of these topological tools in this domain. Simulated experiments using a physics engine and a model of the Baxter robot show that the proposed topology-driven solution is achieving significantly higher success rate in solving cluttered problems in constrained, shelf-like workspaces. The comparison points include randomized baselines as well as alternatives from the literature. The proposed strategy manages to keep the number of pushing actions low, the pushing operations are effective given the physical modeling of the engine, it is computationally efficient and the resulting decisions and motion appear natural for effectively solving such tasks.  

\label{sec:intro}

\section{Related Work}

Many manipulation tasks, such as pick-and-place operations \cite{zeng2018robotic, huang2018case} and object rearrangement \cite{stilman2007manipulation, krontiris2015dealing}, can be solved by using {\bf prehensile actions}, where the robot grasps one object at a time. With relatively predictable movement of the objects, the objective is primarily focused on minimizing the number of grasps to fulfill the task \cite{han2018complexity, wang2021uniform, labbe2020monte}, or maximizing the number of objects to pick within a given time limit \cite{han2019toward}. Prehensile manipulation, however, may require good knowledge of the objects' 3D shape or pose, which can be challenging in cluttered setups. Increasing attention has been paid recently to utilizing machine learning to generate high-quality grasps \cite{varley2017shape, mousavian20196} even without perfect models and accurate pose estimates. Still, however, picking and placing individual objects one at a time can be slow. 

For manipulation setups where it is difficult or slow to perform grasping, {\bf non-prehensile actions} are used for reconfiguring multiple objects at a time, which enables large-scale object manipulation \cite{huang2019large, song2020multi}. Pushing actions are preferred in tasks, such as bin picking and sorting \cite{nieuwenhuisen2013mobile, shome2019towards, song2019object}, as they can be performed with minimalistic end-effectors that can easily fit in a cluttered, constrained space. In harder problems, pushing and grasping actions are used interchangeably throughout the task \cite{pan2020decision, havur2014geometric}. As the effect of pushing is relatively less predictable, there are efforts that focus on better estimating the outcome of pushing actions \cite{zhou2019pushing, zhou2018convex, huang2021dipn}. This paper utilizes pushing actions and performs topological reasoning to identify a group of objects, which can be pushed simultaneously so as to clear the path for approaching a target object to be retrieved. 

Prior efforts in {\bf object retrieval} under clutter have focused on identifying which objects to relocate so as to enable a collision-free path to reach the target object \cite{lee2019efficient, nam2019planning}. Uncertainty arising from occlusion or sensor noise complicate this problem and efforts aim to minimize their effects by inferring object shape \cite{price2019inferring, wong2013manipulation}, reasoning about pose uncertainty \cite{wang2020safe}, or probabilistic filtering \cite{poon2019probabilistic}. To improve the success rate of such tasks, human interaction has been considered to supply a high-level plan, which informs an ordered sequence of objects and approximated goal positions \cite{papallas2020non}. A fast kinodynamic planner that takes advantage of dynamic, nonprehensile actions has been proposed without the assumption of quasi-static interactions \cite{haustein2015kinodynamic}. The method proposed in this paper utilizes persistent homology to improve the efficiency of pushing, which outperforms some of these prior efforts \cite{papallas2020non, haustein2015kinodynamic} in terms of the number of pushing actions needed and planning time. A similar problem relates to pushing a target to the desired goal position \cite{cosgun2011push}, which has been approached via learning an optimal policy from visual input \cite{yuan2018rearrangement}, computing effective manipulation states and actions \cite{haustein2019learning}, or a value function based heuristic \cite{bejjani2018planning}.

Topological reasoning has been more frequently applied to robotics applications, due to the high-dimensionality of the configuration space for high degrees of freedom (DoF) manipulators. Often the configuration space is defined as a manifold, which can then be used to study the kinematic constraints \cite{jaillet2012path} or inverse kinematics of redundant manipulators \cite{burdick1989inverse}. Constraint manifold is introduced to plan feasible paths in configuration spaces with multiple constraints \cite{berenson2009manipulation, qureshi2020neural, englert2020sampling, fernandez2020learning}. On the other side, topological techniques can also be deployed to verify path non-existences \cite{basch2001disconnection, mccarthy2012proving}. Persistent homology is used to classify a class of trajectories with varying task-specific properties such as free of obstacles over the largest range of threshold \cite{bhattacharya2015persistent} or path-connectedness \cite{pokorny2016topological}. Here the work utilizes persistent homology to automatically identify connected components of blocking objects so as to maximize the pushing efficiency.

\label{sec:related}

\section{Problem Setup and Notation}
\label{section: problem setup}

Let $\workspace \subset \mathbb{R}^2$ be a bounded polygonal region, where a set of uniformly-shaped cylindrical {\it movable  obstacles} $\objects=\{o_1, \ldots, o_n\}$ and a similarly shaped target object $\target$ reside. A robotic arm is assumed to be able to reach with its gripper $g$ objects at any position $p \in \workspace$. The arm is not able to pick objects with overhand grasps or lift them due to limited accessibility. The robot arm may access the interior of $\workspace$ from one edge of $\partial\workspace$. Collisions between the arm and the boundary of $\workspace$ beyond that edge are not allowed. In other words, the robot arm moves along a 2D plane where the objects are located while respecting the geometric constraints of the workspace.  The robot geometry assumed by the methods corresponds to a gripper $g$ attached to a cylindrical link representing the rest of the arm's geometry. Given the gripper's pose $s$, the location of the link attached to the gripper is fully specified. For the arm to be able to reach the target $\target$, the gripper must be able to grasp $\target$ and the cylindrical link of the arm must be collision free with all of the movable obstacles and the workspace boundary. In the accompanying experiments, the robot is actually a 7-dim. articulated Baxter robotic arm. The part of the arm inside the workspace is always contained by a cylindrical approximation considered by the proposed methods. The objective is for the robotic arm to reach and grasp the target $\target$, which may require moving some of the obstacles in $\objects$.

The configuration $\config$ of this problem is defined as $\config = \{ \positions, p_{\target}, s\} = \{ (p_1, \ldots, p_n), p_{\target}, s \} \in \workspace^{n+1}\times SE(2)$, where $p_i \in \workspace$ defines the position of $o_i$, i.e., the coordinates of $o_i$'s centroid, $p_{\target}$ defines the position of $\target$, and $s$ is the position and orientation of the robot's gripper $g$. $\config[o_i]=p_i$ indicates that the object $o_i$ assumes the position $p_i$, $\config[\target]=p$ means that $\target$ is at position $p$ while $\config[g]=s$ indicates that the gripper $g$ of the robot is at pose $s \in SE(2)$. Define as $V(p)$ the subset of $\workspace$ occupied by an object at position $p$. A configuration $\config$ is \emph{feasible} if no object-object collisions occurs, i.e., $\config$ is feasible if $\forall i,j \in [1,n], i \neq j: V(\config[o_i]) \cap V(\config[o_j]) = \emptyset$ and $\forall i \in [1,n]: V(\config[o_i]) \cap V(\config[\target]) = \emptyset$.

\begin{wrapfigure}[11]{l}{0.2\textwidth}
    \centering
    \begin{overpic}[width=0.2\textwidth]{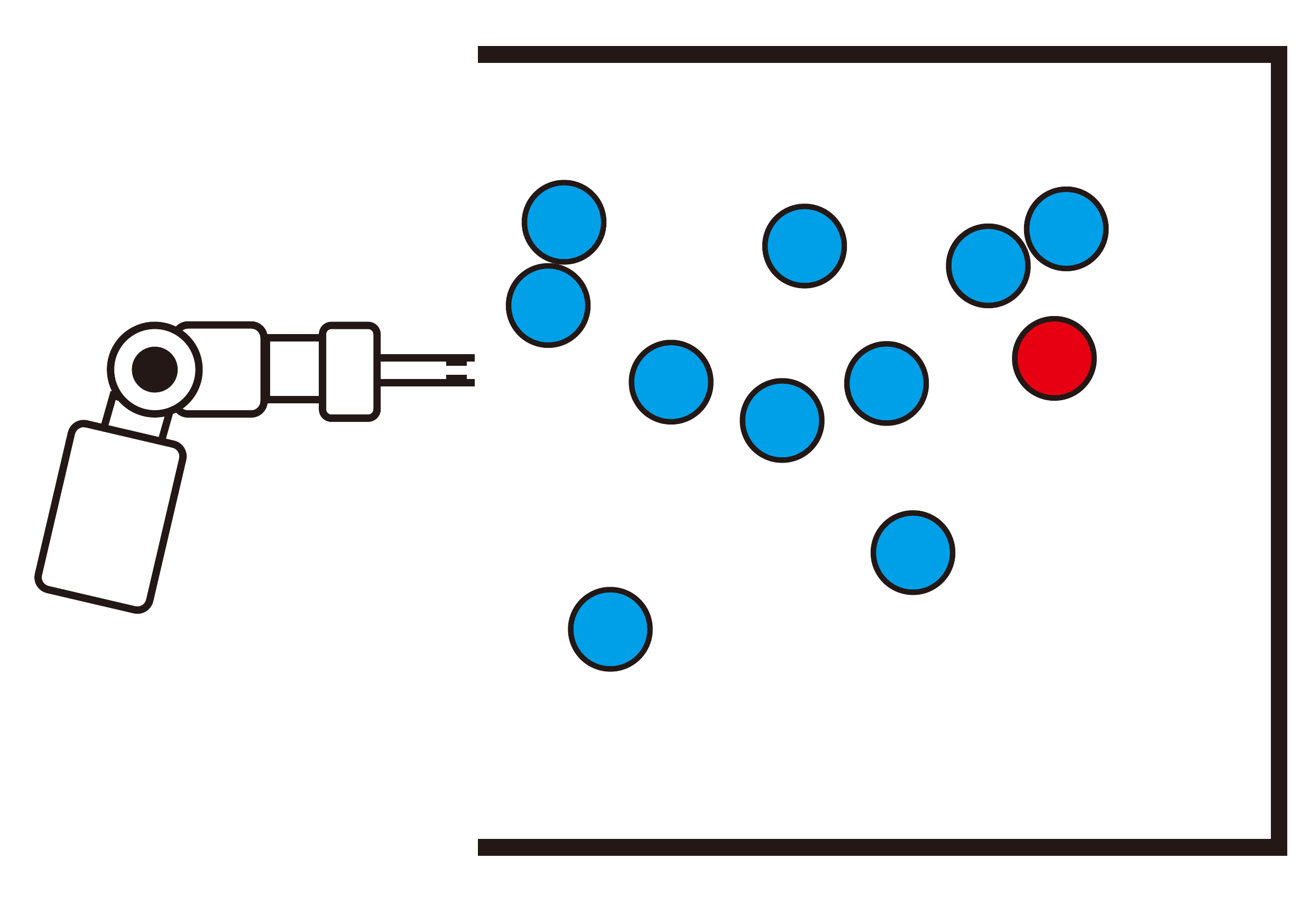}
    \put(80,33){\small\target}
    \end{overpic}
    \vspace{-.25in}
    \caption{Illustration of the workspace $\workspace$ and the robot arm for the same configuration as in Fig. \ref{fig:setup}, where $\target$ is the target object.}
    \label{fig:2dworkspace}
\end{wrapfigure}
Given a feasible configuration $\config$, the arm can push multiple objects at a time. For an object $i$ in the set $I \subset \objects$ of objects affected by the push, the object is relocated from its current position $p_i = \config[o_i]$ to a new position $p^{\prime}_i$, giving rise to a new configuration $\config^{\prime}$, where $\config^{\prime}[o_i] = p^{\prime}_i$ and $\forall j \in [1,n], o_j \notin I: \config^{\prime}[o_j] = \config[o_j]$. The arm's motion results in continuous paths $\pi: [0,1] \to \workspace$ with $\pi(0) = c$ and $\pi(1) = c'$, where $c$ and $c'$ are positions in $\workspace$ obtained by the methods described below.

% \begin{figure}[t]
%     \centering
%     \begin{overpic}[width=0.2\textwidth]{Persistent Homology applied to high-level planning for pushing actions/figures/TwoDExample1_raw.pdf}
%     \put(80,33){\small\target}
%     \end{overpic}
%     % \includegraphics[width=0.2\textwidth]{Persistent Homology applied to high-level planning for pushing actions/figures/TwoDExample1_raw.pdf}
%     \caption{Illustration of the workspace $\workspace$ and the robot arm for the same configuration as in Fig. \ref{fig:setup}, where $\target$ is the target object.}
%     \label{fig:2dworkspace}
% \end{figure}

\begin{figure}[ht]
    \begin{center}
\begin{overpic}[width=\columnwidth]{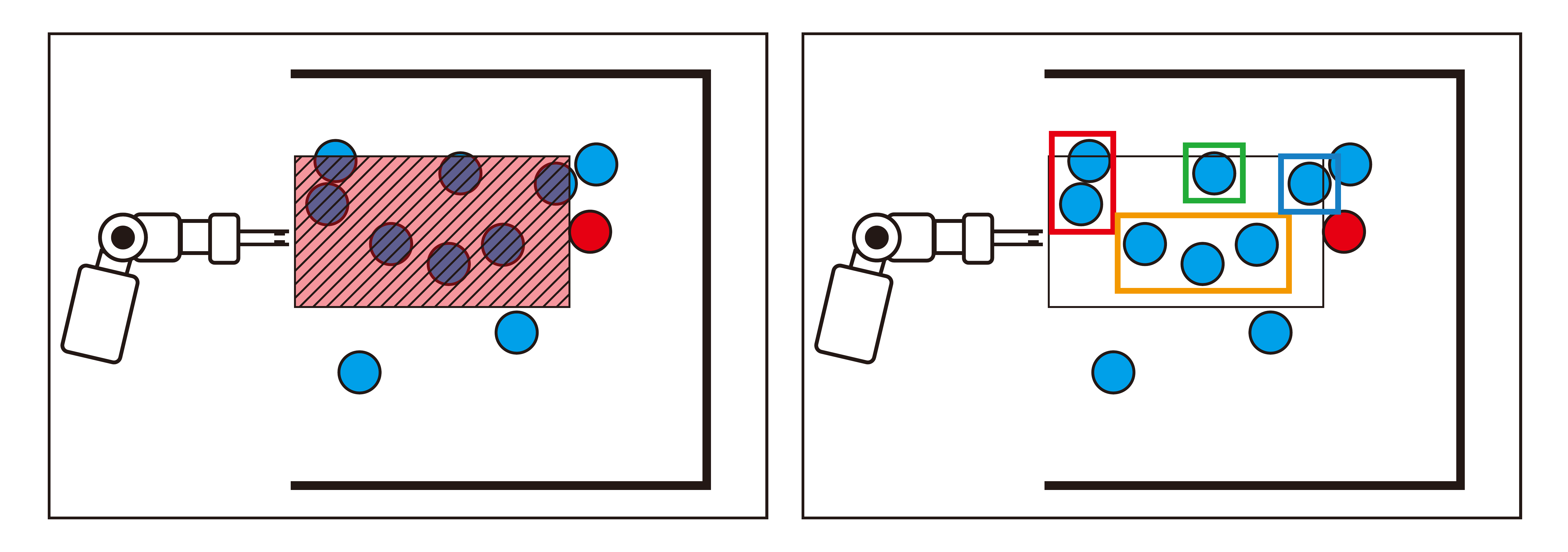}
\put(15, 30){\small $N$}
\put(15, 3){\small $S$}
\put(12, 12){\small $\pathregion(\config)$}
\put(59, 26){\small \textcolor{red}{$CC_c$}}
\put(60, 12){\small $\pathregion(\config)$}
\end{overpic}
\end{center}
\vspace{-3mm}
\caption{\label{fig:pathregion}(left) The path region $\pathregion(\config)$. (right) Connected components in  $\pathregion(\config)$ using $r = 0.086$, where $CC_c = CC_c(\config, r)$ is the closest connected component to the gripper.}

\end{figure}

\begin{figure}[ht]
    \begin{center}
\begin{overpic}[width=\columnwidth]{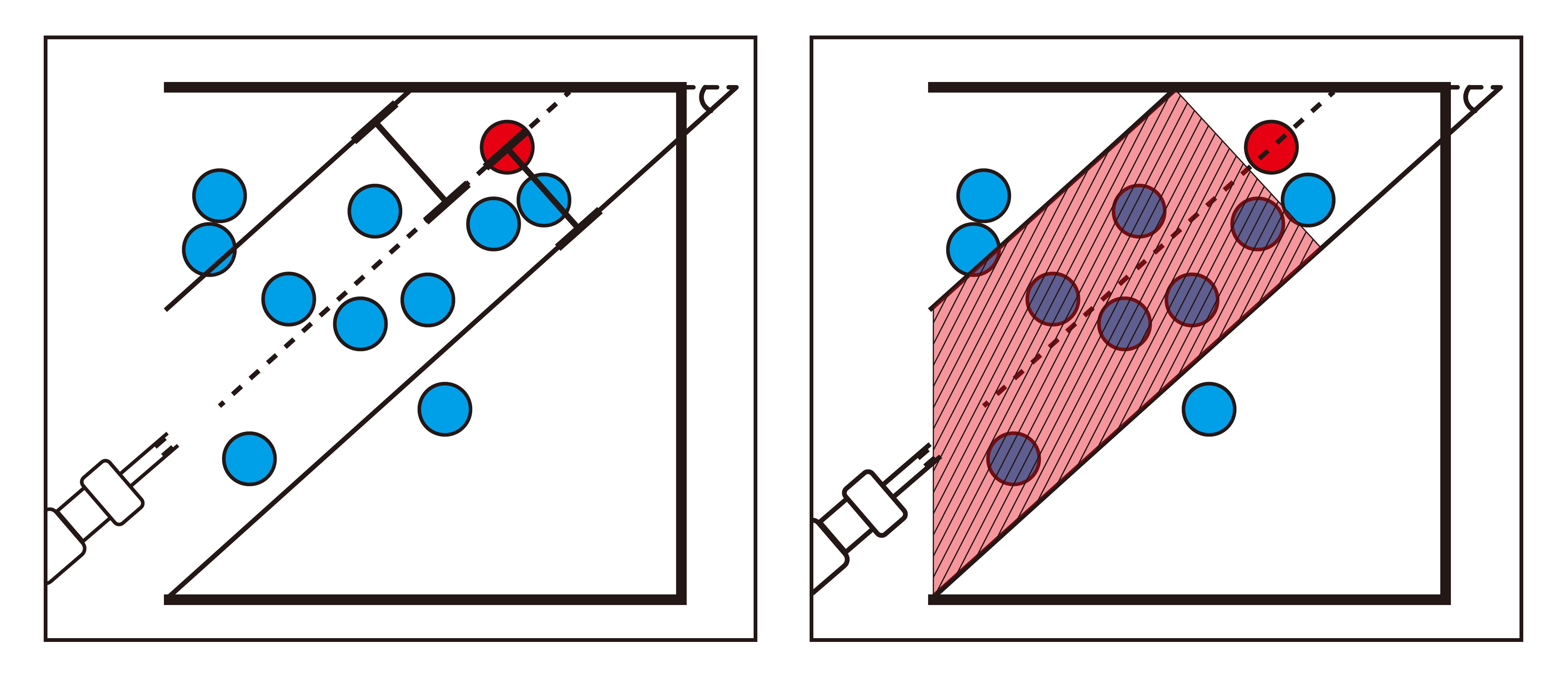}
\put(45, 33){$\phi$}
\put(27, 34){$\omega$}
\put(37, 34){$\omega$}
\put(94, 33){$\phi$}
\put(71, 12){\small $\pathregion(\config)$}
\put(22, 13){$\ell$}
\end{overpic}
\end{center}
\vspace{-3mm}
\caption{\label{fig:phi} Path region $\pathregion(\config)$ with an incidence angle $\phi$, where $w$ is the width of the arm and $\ell$ is the straight line, which passes through $(0,0)$ and below $\config(\target)=p$ with $d(p,\ell)=w$.}

\end{figure}

\section{Method}
\label{sec:method}

%\subsection{Foundations of Persistent Homology}

Persistent homology is a tool from algebraic topology that computes topological features of a space at different spatial resolutions. It has been extensively applied for topological analysis of point-cloud data (e.g., see \cite{MR3328629} and \cite{KaczynskiTomasz2004Ch/T}). In summary, given a collection of points, consider growing balls of radius $r$ centered at each point. As the balls expand, track the unions of all these balls as they overlap each other given a $1$-parameter family of spaces. For each radius $r$, one can build a Vietoris–Rips complex (abstract simplicial complex) using the information given by the intersection of the $r$-balls (for more details see \cite{KaczynskiTomasz2004Ch/T}). In this setting, persistent homology is the homology of the Vietoris–Rips complex as a function of $r$. Intuitively, persistent homology counts the number of connected components and holes of various dimensions and keeps track of how they change with parameters.

\begin{figure}[t]
\begin{overpic}[width=0.25\textwidth]{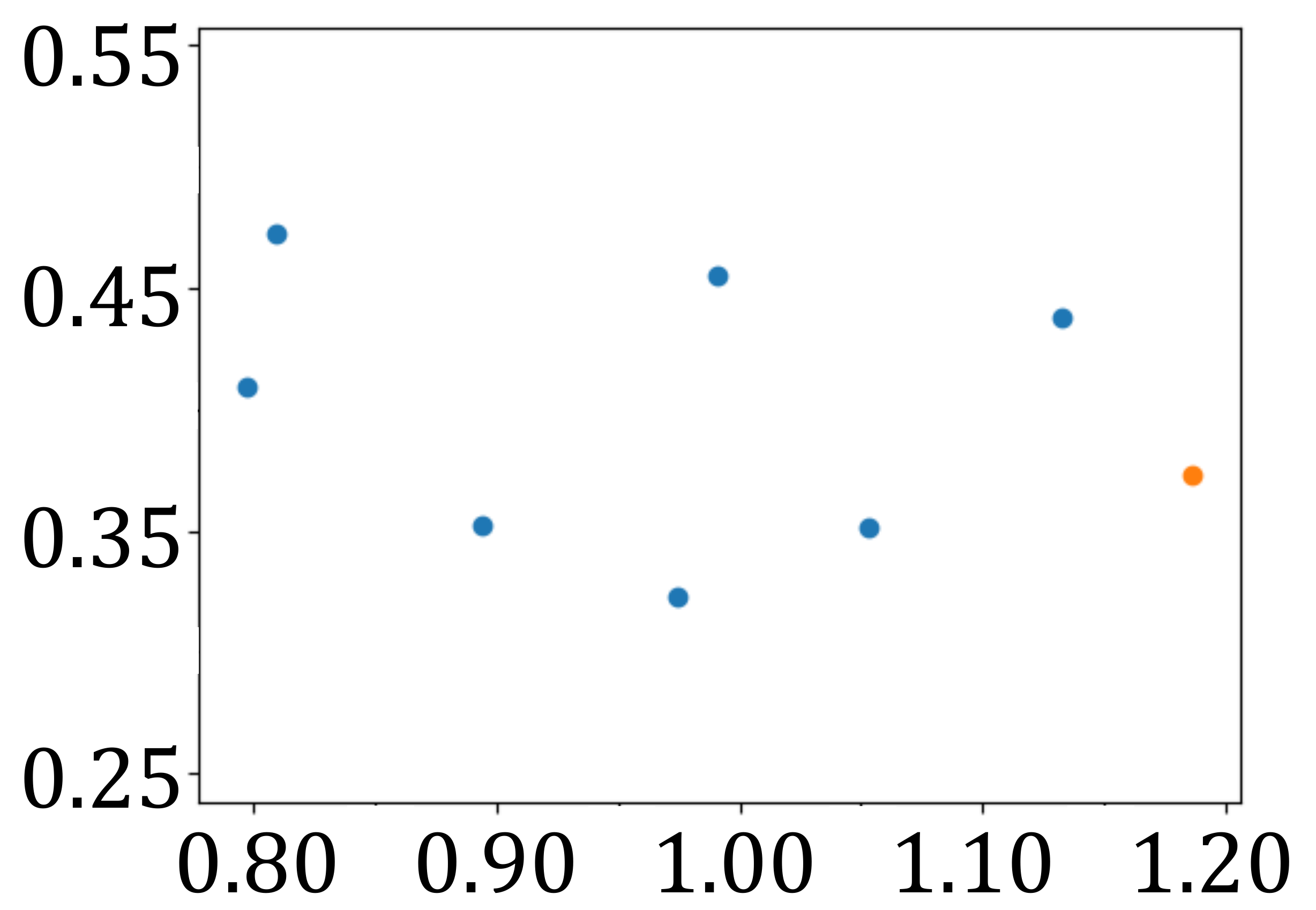}
\put(88, 24){\small $\mathcal T$}
\end{overpic}
\begin{overpic}[width=0.20\textwidth]{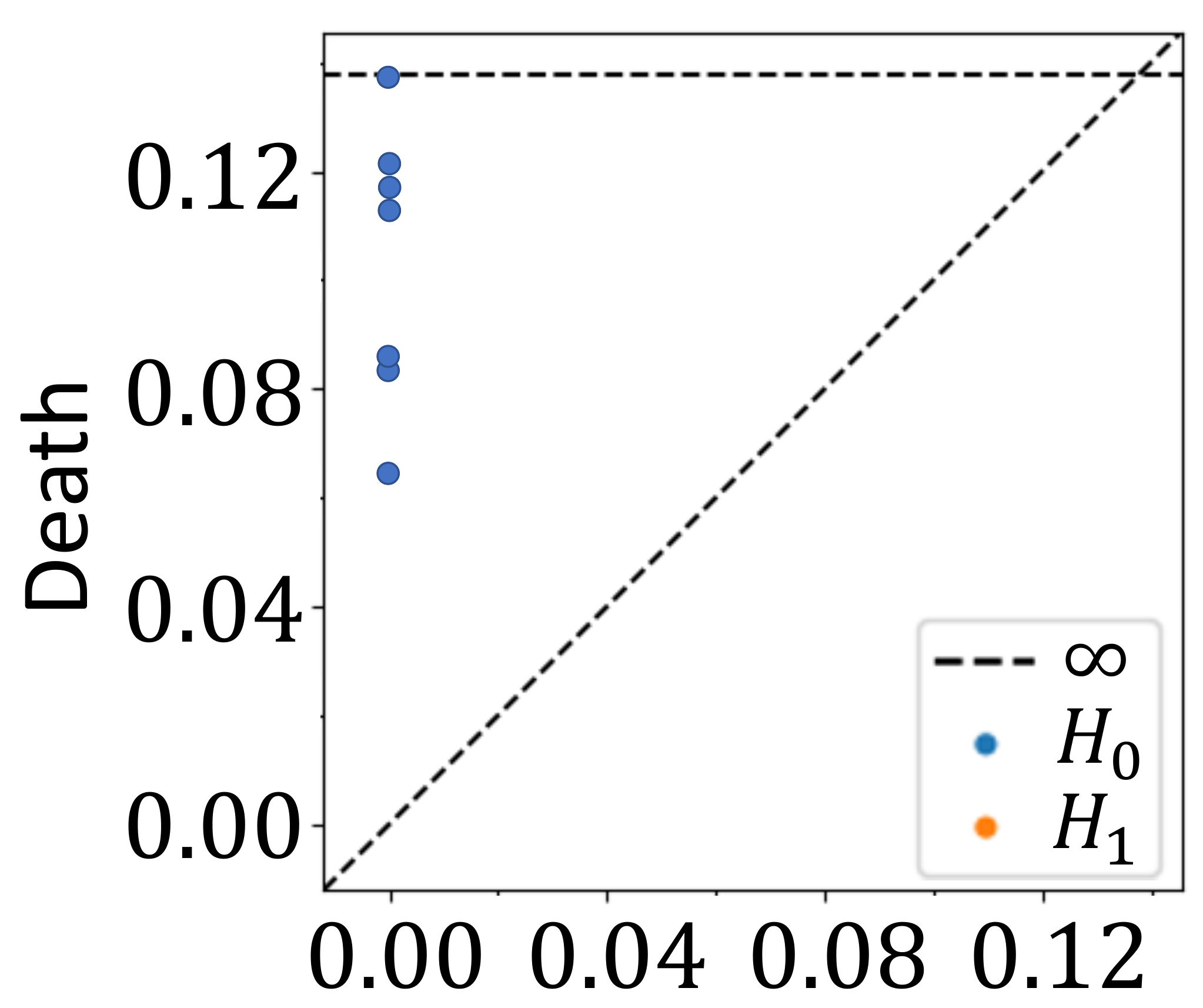}
\end{overpic}
\hspace{-0.6cm}
\caption{(left) The subset of objects in the path region $\pathregion(\config)$ for the problem in Fig. \ref{fig:setup}. Blue ones are obstacles and the orange one (rightmost point) is the target $\target$. (right) The persistence diagram for the points shown in the left figure.}
    \label{fig:pers_diagram}
\end{figure}

\begin{figure}[t]
    \centering
    \includegraphics[width=0.46\textwidth]{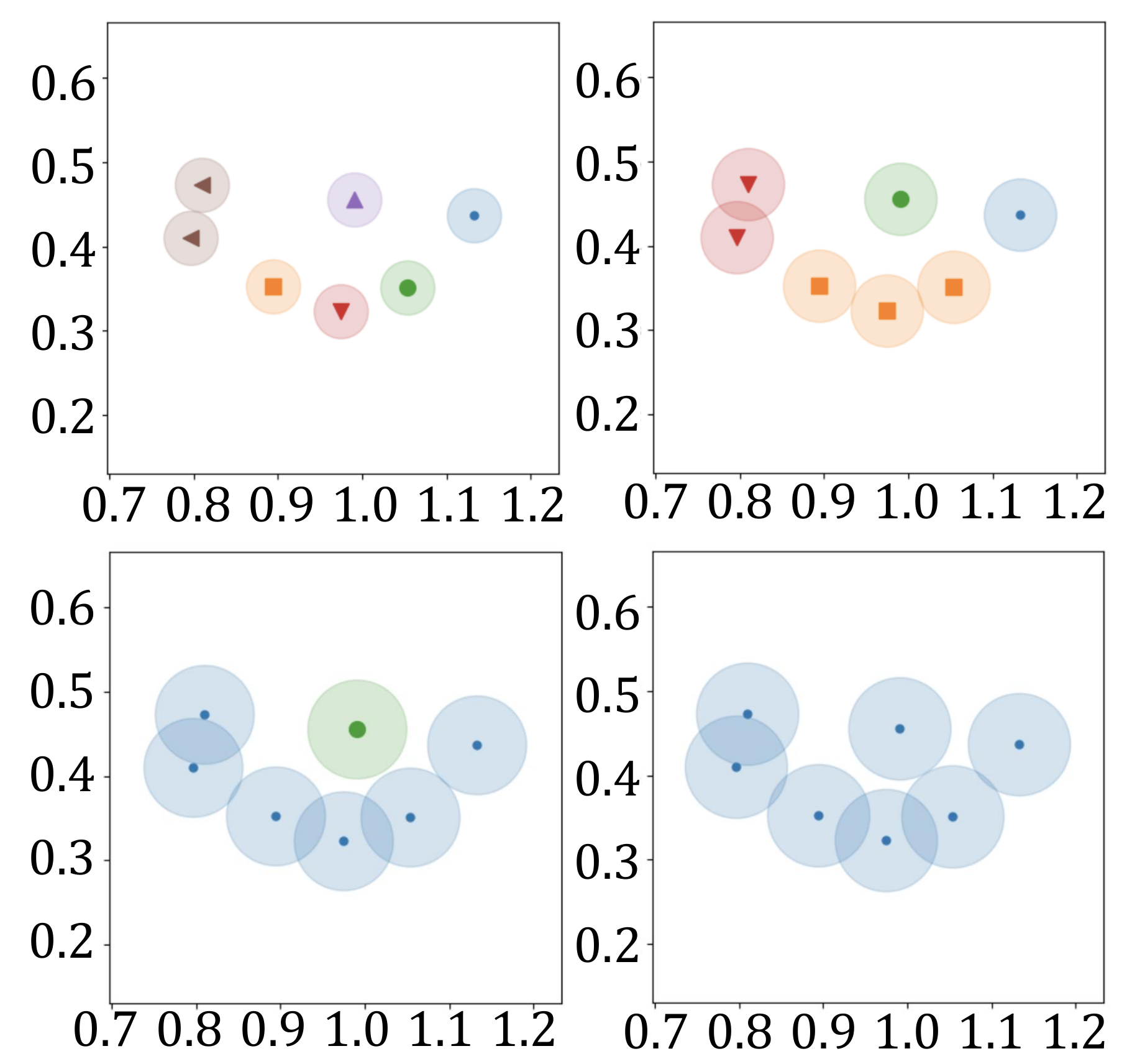}
    \caption{Examples of connected components that persist obtained for four different radii $r$ in the persistence diagram in Fig. \ref{fig:pers_diagram}. (upper-left) For $r = 0.064$, there are 6 different connected components shown by different markers.  (upper-right) For $r = 0.086$, there are 4 connected components. (bottom-left) For $r = 0.121$, two connected components. (bottom-right) Only one connected component (it contains all points) for $r \geq 0.14$.}
    \label{fig:connec_comp}
\end{figure}

Since we are interested in computing connected components that persist, we focus on the zeroth-homology (zero dimensional homology) and use the field $\mathbb{Z}_2$ as a coefficient for the homology. As the radius $r$ increases, the zero dimensional persistent homology records when the ball in one connected component first intersects a ball of a different connected component, merging both connected components in one, see Figs. \ref{fig:pers_diagram} and \ref{fig:connec_comp}.

\subsection{Proposed Approach for Non-Prehensile Manipulation}

Given a target $\target\in \objects$, we can define a region $\pathregion(\config) \subset \workspace$ smaller than the workspace where we use persistent homology to find connected components and perform pushing actions. We denote $\pathregion(\config)$ as the {\it path region}, where the robot arm will push the obstacles to clean the path to the target. The shape and size of $\pathregion(\config)$ depends on the boundary of $\workspace$ and the width $w$ of the robot arm. Since $\workspace$ is rectangular shelf, hence it is enough to use  $N \subset \workspace$ and $S\subset \workspace$ (the parallel walls of the shelf) to describe $\pathregion(\config)$ by using Algorithm \ref{alg:path_region} and the ideas shown in Figs. \ref{fig:pathregion} and \ref{fig:phi}.

Fix an orientation of the shelf where the axis $x$ corresponds to the depth of the shelf and the axis $y$ is aligned with the width of shelf with $S$ in $y=0$ and $N$ in $y = w_s$, where $w_s$ is the width of the shelf. When the Euclidean distance between $\config(\target) = p$ and $S\cup N$ is greater than $w$, $d(p, S\cup N)> w$, $\pathregion(\config)$ is a rectangle given by the points $(0, \config[\target]_y - w)$ and $(\config[\target]_x, \config[\target]_y + w)$, where $\config[\target] = (\config[\target]_x, \config[\target]_y)$. See Fig. \ref{fig:pathregion} for an illustration.

\begin{figure}[t]
\vspace{1mm}
\centering
    \includegraphics[width=0.23\textwidth]{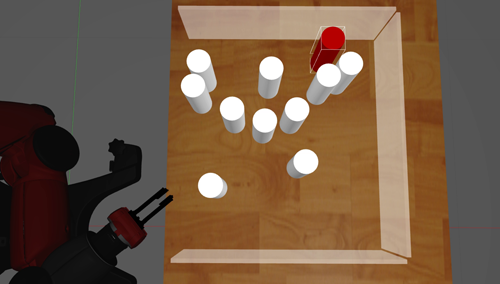}
    \includegraphics[width=0.23\textwidth]{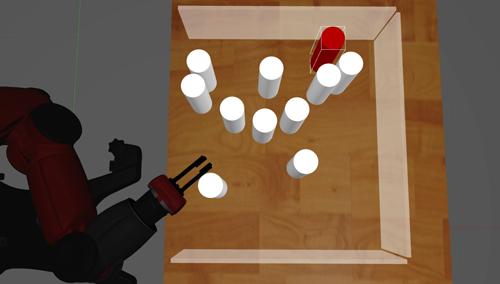}
    \includegraphics[width=0.23\textwidth]{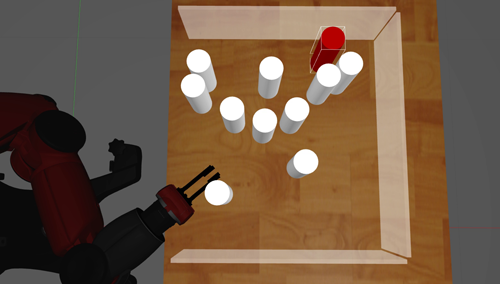}
    \includegraphics[width=0.23\textwidth]{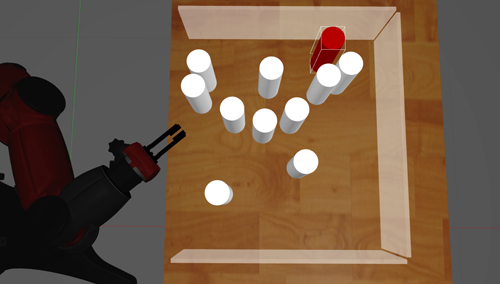}
    \includegraphics[width=0.23\textwidth]{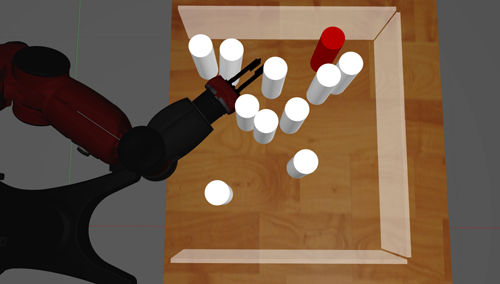}
    \includegraphics[width=0.23\textwidth]{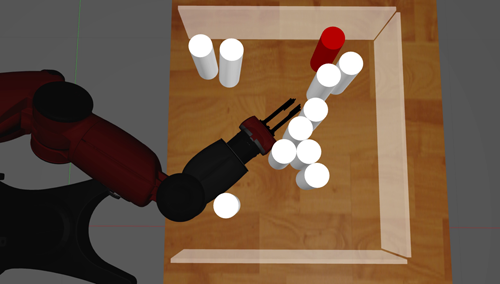}
    \includegraphics[width=0.23\textwidth]{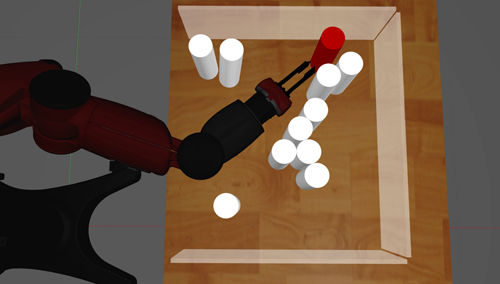}
    \includegraphics[width=0.23\textwidth]{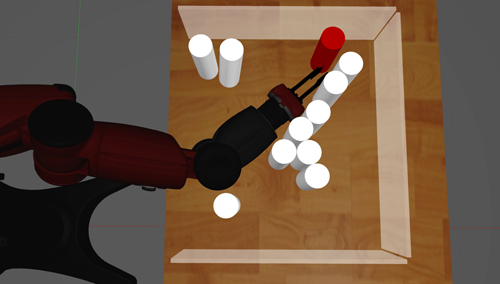}
    \caption{Sequence of pushing actions with angle $\phi = 0.709$, where the selected radius for the first action is $r_m = 0.091$ with $\nu = 0.015$ and $h = 0.08$.}
    \label{fig:pushing actions_phi}
\vspace{-1mm}\end{figure}

For the case where $d(p, S\cup N) \leq w$, the path region $\pathregion(\config)$ has to have an incidence angle $\phi$ with respect to $S$ or $N$, the closest to $p$. Without loss of generality, assume that $d(p, N)< w$, then $\phi$ is the acute angle between $y = w_s$ and straight line $\ell$ that passed through $(0,0)$ with $d(p, \ell) = w$, such that $\config[\target]$ is above $\ell$. See Figs. \ref{fig:phi} and \ref{fig:pushing actions_phi}. We use the planar rotation matrix $R_{-\phi}$ to rotate $\workspace$ to obtain a rotated path region $R_{-\phi}(\pathregion(\config))$, such that it is analogous to the rectangle found for the case $d(p, S\cup N)> w$.

\begin{algorithm}
\caption{Path Region($\config$)}
\label{alg:path_region}
\small
\DontPrintSemicolon
$\pathregion(\config) \gets \emptyset$

\If{$d(p, S\cup N) > w $}{
    \For{$o\in\objects_b$}{
        \If{$ \config[\target]_y - w \leq\config[o]_y \leq \config[\target]_y + w$ \text{ and } $ (\config[g])_x \leq\config[o]_x < \config[\target]_x$}{
            $\pathregion(\config) \gets o$
        }
    }
}
\Else{
    \For{$o\in\objects_b$}{
        \If{$ (R_{-\phi}\config[\target]^T)_y - w \leq (R_{-\phi}\config[o]^T)_y \leq (R_{-\phi}\config[\target]^T)_y + w$ \text{ and } $ (R_{-\phi}\config[g]^T)_x \leq (R_{-\phi}\config[o]^T)_x < (R_{-\phi}\config[\target]^T)_x$}{
            $\pathregion(\config) \gets o$
        }
    }   
}
\KwRet{$\pathregion(\config)$}
\end{algorithm}

Let $\connectedcomp(\config, r)$ be the collection of connected components for the radius $r>0$ inside of $\pathregion(\config)$. We use the software Ripser \cite{ctralie2018ripser} to find the zero dimensional persistent homology, and then to extract the connected component and its generators. For our setting, we only need the closest connected component, $\connectedcomp(\config, r)_c$, to the position of the gripper of the robot arm $\config[g]$, since the arm will try to push all obstacles in $\connectedcomp(\config, r)_c$ to the outside of $\pathregion(\config)$. Such manipulation of objects is non-prehensile, so it is not guaranteed that the new arrangement $\config$ of $\objects$ will preserve all connected components in $\pathregion(\config)$. 

Now, we compute the circumscribed rectangle, $\rect(\config, r)$, of $\connectedcomp(\config, r)_c$ (the smallest rectangle that contains $\connectedcomp(\config, r)_c$, such that the sides of $\rect(\config, r)$ are parallel to the $x$ and $y$ axes). We use it to plan the pushing actions to move all obstacles $\connectedcomp(\config, r)_c$ with one single action. Algorithm \ref{alg:circ_rect} CRCCC (Circumscribed Rectangle of the Closest Connected Components) describes this procedure. 

\begin{algorithm}
\caption{CRCCC($\config, r$)}
\label{alg:circ_rect}
\small
\DontPrintSemicolon
$\pathregion(\config) \gets $ \FuncSty{PathRegion}($\config$)  

$\connectedcomp(\config, r) \gets$ \FuncSty{Ripser}($\pathregion(\config)$, $r$, dim $= 0$)

$\connectedcomp(\config, r)_c \gets$ \FuncSty{ClosestSet}($\connectedcomp(\config, r)$, $s$)

$\rect(\config, r) \gets$ \FuncSty{CircumscribedRectangle}($\connectedcomp(\config, r)_c$)

\KwRet{$\rect(\config, r)$}
\end{algorithm}

With the circumscribed rectangle of the closest connected components $\rect(\config, r)$ the arm will push the region described by $\rect(\config, r)$ either to up or down. It depends on the location of $\rect(\config, r)$ relative to the path region $\pathregion(\config)$, if the $y$ coordinate of the centroid of $\rect(\config, r)$, $c_{\rect(\config, r)y}$, is higher than the $y$ coordinate of the centroid of $\pathregion(\config)$, $c_{\pathregion(\config)y}$. This will decide whether the arm will do a sweeping movement from the bottom of $\rect(\config, r)$ to top until the arm reaches $N$, the north boundary of $\pathregion(\config)$. The opposite is executed if $c_{\rect(\config, r)y} < c_{\pathregion(\config)y}$ . For the case where $c_{\rect(\config, r)y} = c_{\pathregion(\config)y}$ perform the same analysis with $c_{\rect(\config, r)y}$ and the $y$ coordinate of the centroid of $\workspace$, $c_{\workspace y}$. See Algorithm \ref{alg:push} for details.

\begin{algorithm}
\caption{Pushing Actions($\config, r$)}
\label{alg:push}
\small
\DontPrintSemicolon
$\rect(\config, r) \gets $ \FuncSty{CRCCC}($\config, r$)\;
\eIf{$c_{\rect(\config, r)y}>c_{\pathregion(\config)y}$}{
    \FuncSty{SweepingBottomToTop()}\;
    }{
    \eIf{$c_{\rect(\config, r)y}<c_{\pathregion(\config)y}$}{
        \FuncSty{SweepingTopToBottom()}\;
        }{
        \eIf{$c_{\rect(\config, r)y}\leq c_{\workspace y}$}{
            \FuncSty{SweepingTopToBottom()}\;
            }{
            \FuncSty{SweepingBottomToTop()}\;
            }
        }
    }
$\config \gets \FuncSty{UpdateConfiguration}()$\;
$t \gets \FuncSty{ActionsTime}()$\;
\KwRet{(\config, t)}
\end{algorithm}

\begin{figure}[t]
\vspace{1mm}
    \centering
    \includegraphics[width=0.23\textwidth]{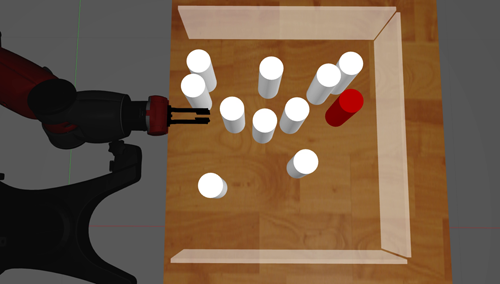}
    \includegraphics[width=0.23\textwidth]{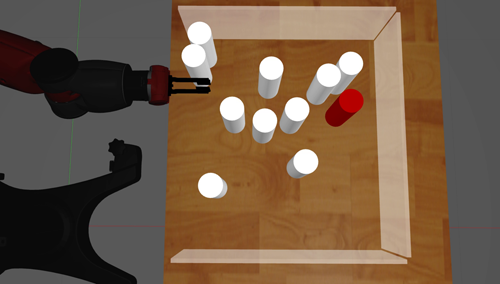}
    \includegraphics[width=0.23\textwidth]{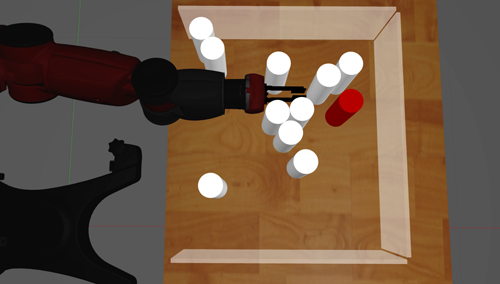}
    \includegraphics[width=0.23\textwidth]{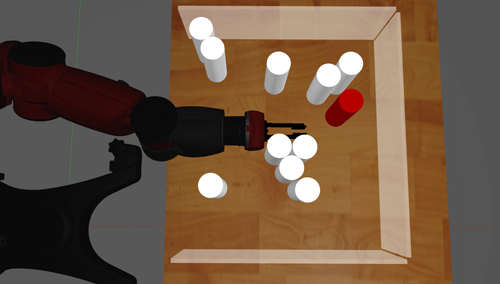}
    \caption{Sequence of pushing actions, where the selected radius $r_m = 0.086$ for the first action is the minimal of $R = R_{\nu, h}$ for $\nu = 0.015$ and $h = 0.08$. Such radius $r_m$ produces the connected components described in Fig. \ref{fig:connec_comp} with $r = 0.086$.}
    \label{fig:pushing actions}
\vspace{-1mm}
\end{figure}

Up this point we have computed everything for a given radius $r$. We use the persistence diagram to select the appropriate $r$ in an informed manner. First, define the {\it persistent radius} for a given value $\nu>0$ as a radius $r$ where one or more connected components die. Between $r$ and $r+\nu$ the number of connected components remains the same, in other words the connected components persist. Let $R_\nu$ be the set of all persistent radii for a given value $\nu>0$. In the persistence diagram in Fig. \ref{fig:pers_diagram}, $R_{0.015}=\{0.064, 0.086, 0.121, 0.14\}$. Note that, $R_h$ is not empty since the radius where the last connected component dies belongs to $R_\nu$.

For our setup, $\nu$ is considered as the margin of error for the pushing action when the arm is moving a whole connected component outside the path region $\pathregion(\config)$. 

Observe that $R_\nu$ may contain small radii such that the gripper cannot move between two connected components, hence we only consider persistent radii greater than $h$, the number given by 110$\%$ of width of the gripper (10$\%$ more to consider imprecision) plus two times the radius of the objects. Define $R(\config)=R_{\nu, h}(\config)$ as the collection of persistent radii greater than $h$ for a configuration space $\config$. For example, in the persistence diagram of Fig. \ref{fig:pers_diagram}, $R_{\nu, h}(\config)=\{0.086, 0.121, 0.14\}$ for $\nu = 0.015$ and $h = 0.08$.

The simplest approach to solve the reaching through clutter problem is to select $r_m$ as the minimum element of $R(\config)$ for each configuration space after each pushing action. See Algorithm \ref{alg:simplest} for the steps.

\begin{algorithm}
\caption{PH Informed Actions($\config_0, \nu, h$)}
\label{alg:simplest}
\small
\DontPrintSemicolon

$\config \gets \config_0$

\While{$\pathregion(\config)$ is not empty}{
$D_0 \gets $ \FuncSty{Ripser}($\pathregion(\config), dim = 0$) 

\hspace{2cm} $\backslash\backslash$ \text{ 0d persistence diagram }

$R(\config) \gets$ \FuncSty{SetOfPersistentRadii}($D_0, \nu, h$)

$r_m \gets $ \FuncSty{MinimumElement}($R(\config)$) 

$\config' \gets$ \FuncSty{PushingActions}($\config, r_m$)

\If{$\config' = \config$}{
    label $\config$ failure
    }
}
\end{algorithm}

Another approach to the problem is to use not only $r_m$ but all radii in $R_{\nu, h}(\config)$ for a configuration space $\config$.  Specifically, it is possible to build a tree with nodes corresponding to configurations in $\config$ and propagate it by performing the push action for each radius in $R_{\nu, h}(\config)$. The propagation is performed until it achieves success or failure.  Finally, it searches for the path that leads to success and minimizes time for performing the pushing actions as in Algorithm \ref{alg:search}.

%Create a tree, where the root is the initial configurations and the leaves are given by the radius obtained by the next interaction (considering only radius where the connected components persist). After simulations, select the path in the tree that is the fastest and successful.

\begin{algorithm}
\caption{PH Informed Search($\config_0, \nu, h$)}
\label{alg:search}
\small
\DontPrintSemicolon
\SetKw{Continue}{continue}
tree $\gets \{ \text{nodes} = \{ \config_0\}, \emptyset \}$\;
$Q \gets \{\config_0\}$\;
\While{$Q \neq \emptyset$}{
        $Q' \gets \emptyset$\;
        \For{$\config \in Q$}{
            \If{$\pathregion(\config) = \emptyset$}{ label $\config$ success \Continue}
            $D_0 \gets $ \FuncSty{Ripser}($\pathregion(\config), dim = 0$)\;
            $R(\config) \gets$ \FuncSty{SetOfPersistentRadii($D_0, \nu, h$)}\;
            \For{$r \in R(\config)$}{
                ($\config_{new}, t_{new}) \gets $ \FuncSty{PushActions($\config, r$)}\;
                \If{$\config = \config_{new}$}{ label $\config$ fail \Continue}
                tree.nodes = tree.nodes$\cup \{\config_{new}\}$\;
                tree.edges = tree.edges$\cup\{(\config, \config_{new}), t\}$\;
                $Q' = Q' \cup \{\config_{new}\}$
                }
            }
        $Q \gets Q'$\;
        }
path $\gets$ \FuncSty{ShortestSuccessfulPath}(tree)
\end{algorithm}

% \begin{algorithm}
% \caption{PH Informed Search($\config_0, \nu, h$)}
% \label{alg:search}
% \small
% \DontPrintSemicolon
% tree $\gets \{ \text{nodes} = \{ \config_0\}, \emptyset \}$\;
% \SetKwFunction{propagate}{Propagate}
% \SetKwProg{Fn}{Function}{:}{}
% \Fn{\propagate{$\config$}}{
%     \eIf{$\pathregion(\config) = \emptyset$}{
%         label $\config$ success
%         }
%         {
%         $D_0 \gets $ \FuncSty{Ripser}($\pathregion(\config), dim = 0$)\;
%         $R(\config) \gets$ \FuncSty{SetOfPersistentRadii($D_0, \nu, h$)}\;
%         \For{$r \in R(\config)$}{
%             ($\config_{new}, t_{new}) \gets $ \FuncSty{PushActions($\config, r$)}\;
%             \eIf{$\config = \config_{new}$}{
%                 label $\config$ failure}{
%                 tree.nodes = tree.nodes$\cup \{\config_{new}\}$\;
%                 tree.edges = tree.edges$\cup\{(\config, \config_{new}), t\}$\;
%                 \FuncSty{Propagate}($\config_{new}$)
%                 }
%             }
%         }
%     }
% \FuncSty{Propagate}($\config_0$)\;
% path $\gets$ \FuncSty{ShortestSuccessfulPath}(tree)
% \end{algorithm}

\section{Experiments}
\label{sec:experiments}

Experiments were run on a Ubuntu workstation with {\it Intel Core i5-8259U 3.8Ghz 16GB RAM}. Gazebo \cite{gazebo} was used for simulating the task execution and MoveIt \cite{moveit} was used for forming motion planning queries to the Bi-EST planner \cite{hsu1997path} in OMPL \cite{ompl}.

For the comparison experiments of the GRTC-Heuristic, a simplified version of the algorithm in \cite{papallas2020non} was used that pushes cylinders in a straight line to their goal region. The straight pushes are generated using a kinematic motion planner. This varies from the original GRTC-Heuristic, which uses a kinodynamic motion planner to randomly sample push actions to get the cylinder into a goal region.

We also run experiments using a version of the algorithm PH Informed Actions \ref{alg:simplest} where we remove the persistent homology information. More specifically, the lines 3-6 are changed by $r_m \gets 0.01$, that is, an algorithm that does not use persistent homology and does not group obstacles. We denote this algorithm by OOA (one by one action). The idea is to compare the statistical difference between non-prehensile manipulation without grouping the obstacles and the PH informed actions.

\begin{figure}[t]
\vspace{1mm}
    \centering
    \includegraphics[width=0.23\textwidth]{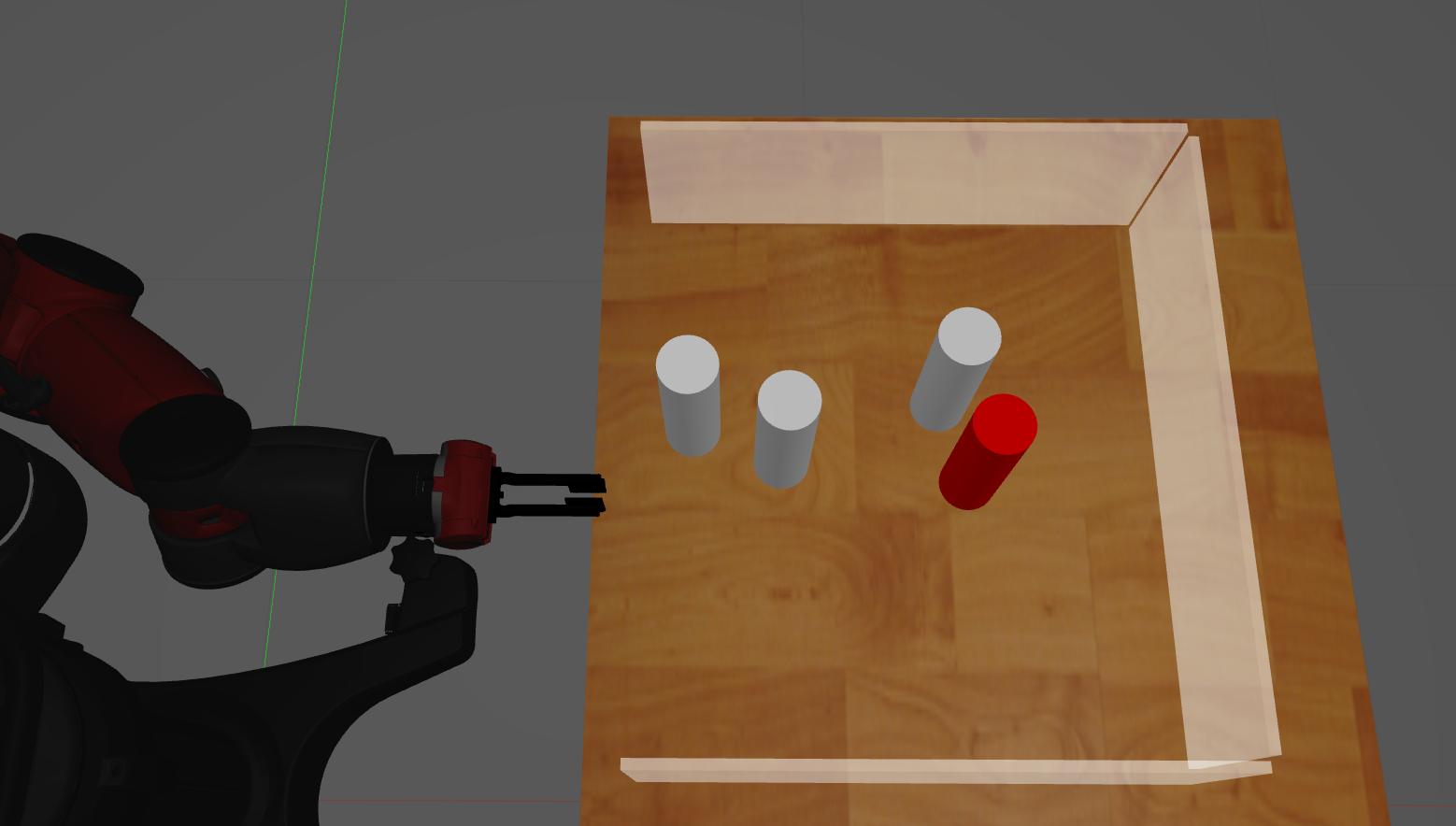}
    \includegraphics[width=0.23\textwidth]{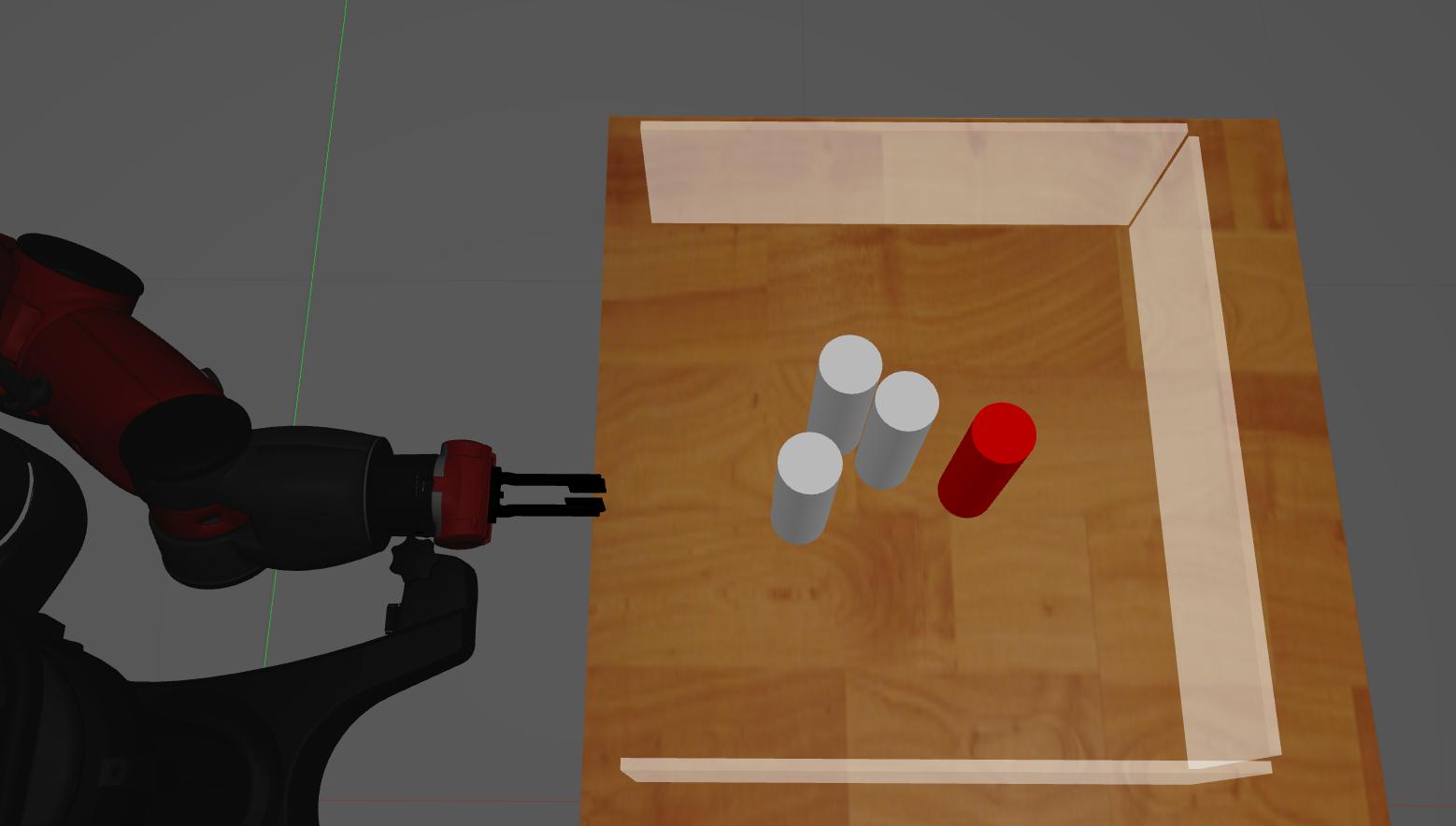}
    \includegraphics[width=0.23\textwidth]{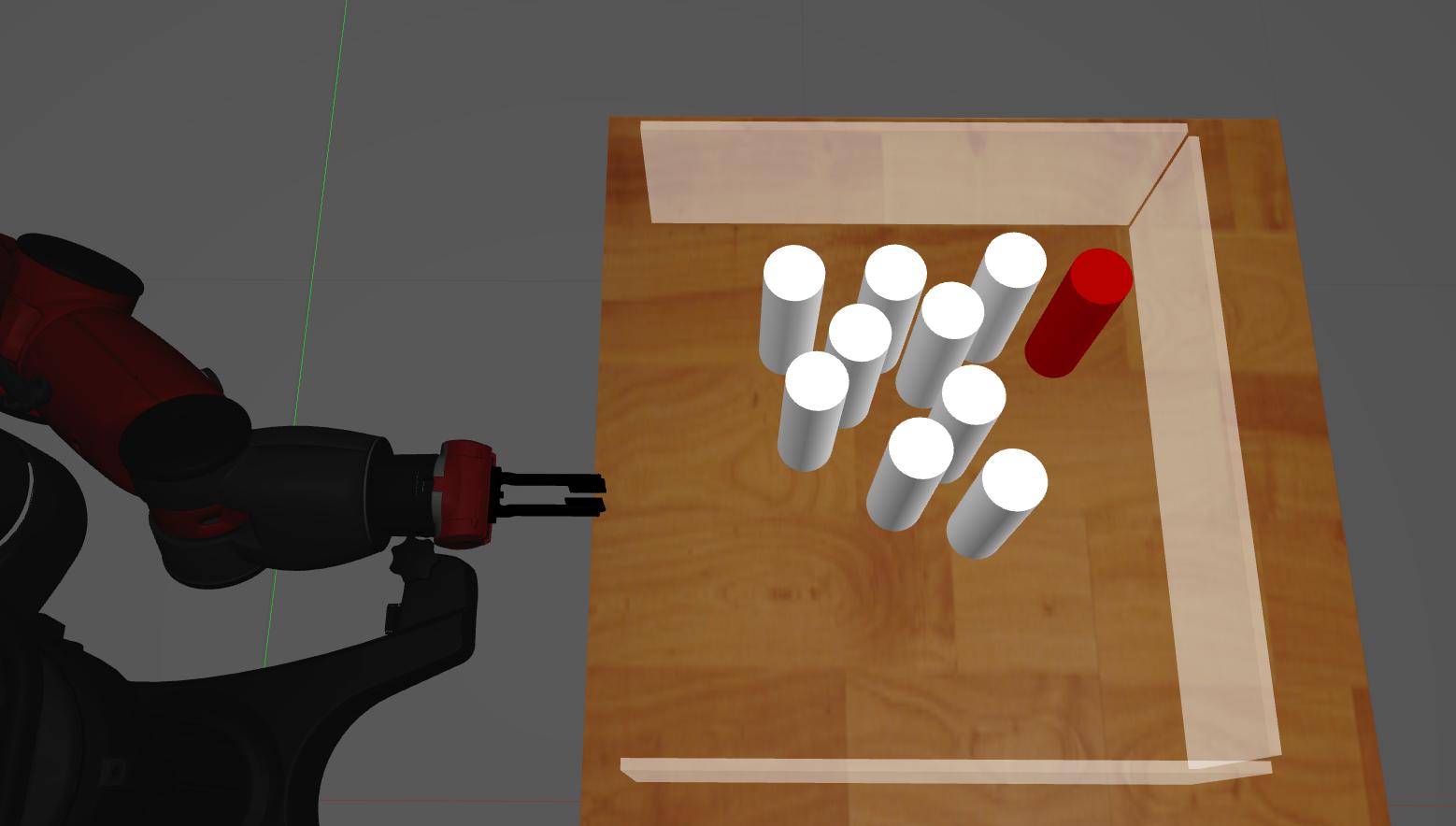}
    \includegraphics[width=0.23\textwidth]{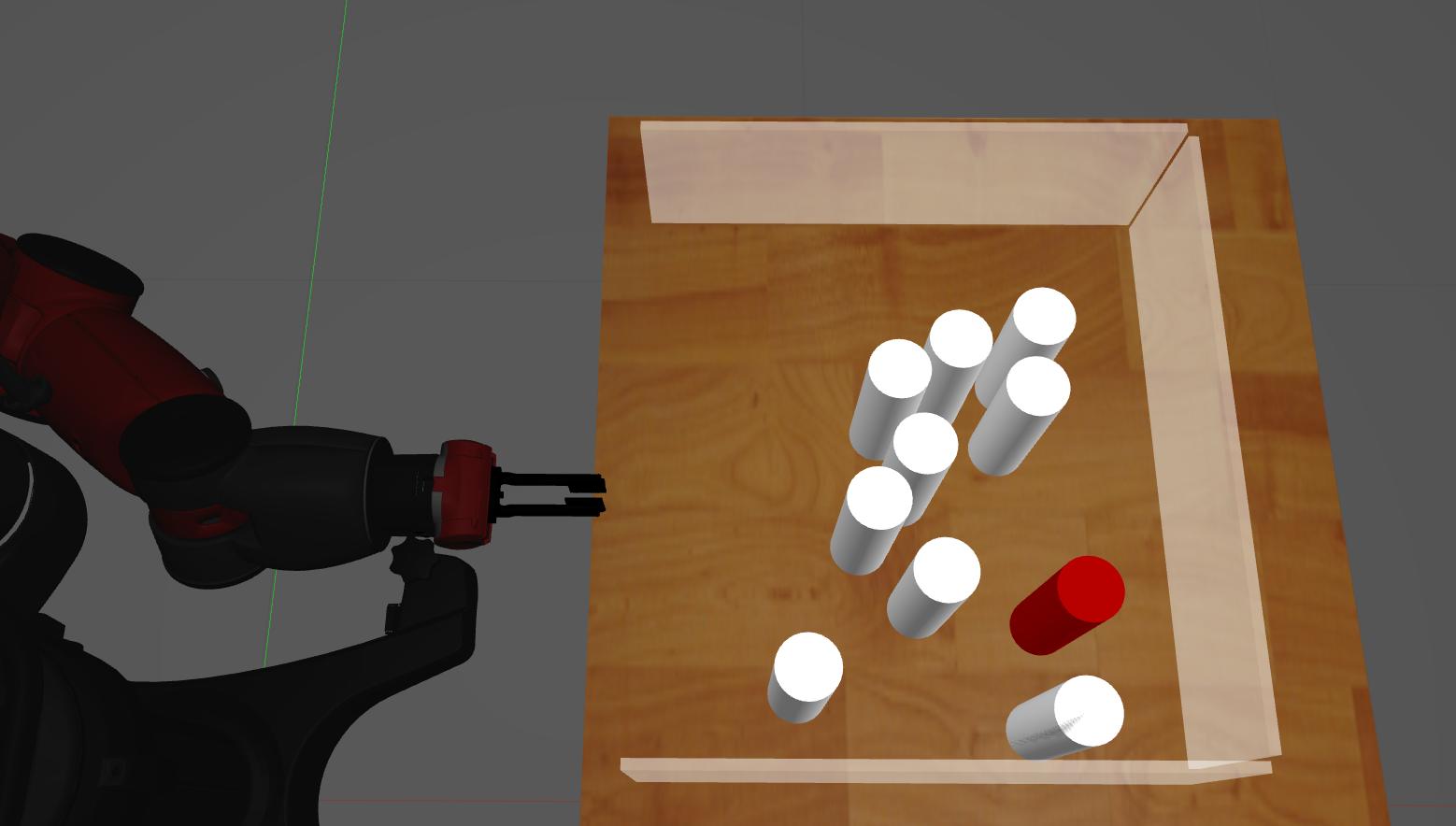}
    \includegraphics[width=0.23\textwidth]{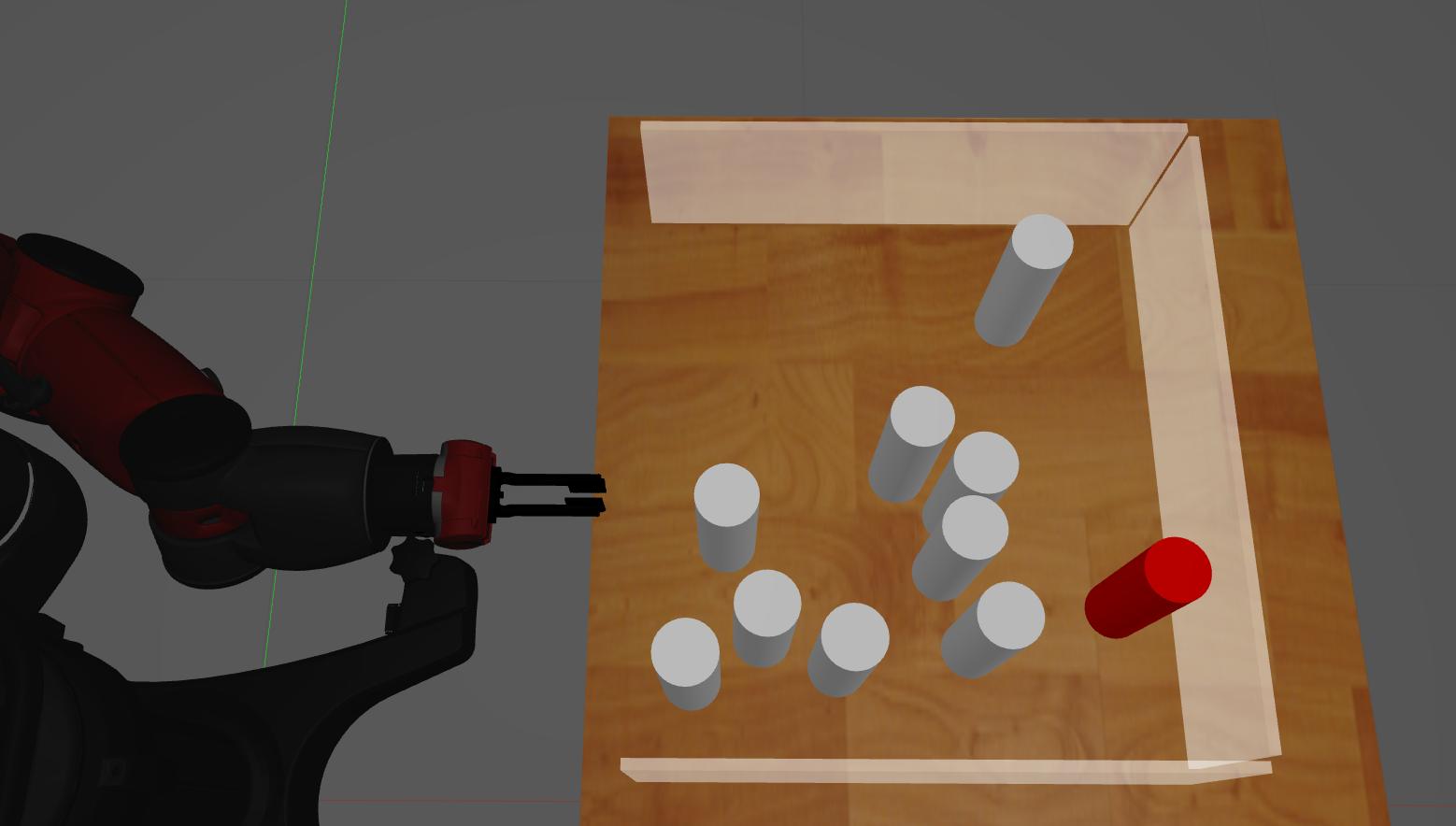}
    \includegraphics[width=0.23\textwidth]{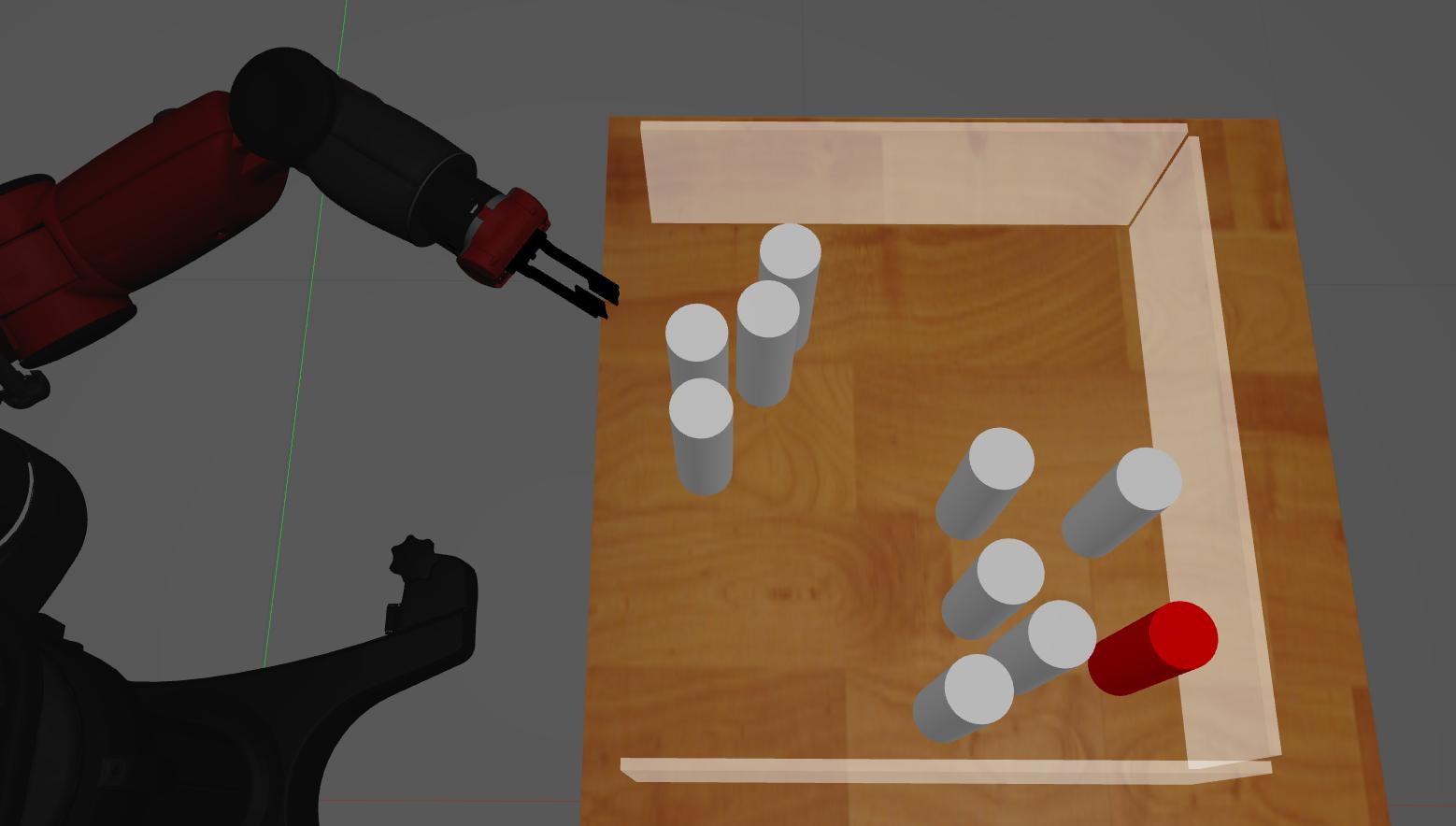}
    \caption{From left to right: (first) simple case number 1; (second) simple case number 7; (third) S1 case from \cite{papallas2020non}; (forth) S7 case from \cite{papallas2020non};
    (fifth) random experiment number 3; (sixth) random experiment number 32. }
    \label{fig:instances}
\vspace{-1mm}    
\end{figure}

For all methods, an overall planning time threshold of 300 seconds was set. If a method exceeds this threshold, it exits and returns a failure to retrieve the assigned object. To evaluate performance of the proposed algorithms we conduct 121 experiments: one manually designed (in Fig. \ref{fig:setup}); 10 instances from a prior work \cite{papallas2020non}; 10 randomly generated instances that are simple (with only 4 objects such that all are far from the wall); and 100 random instances where the target object is always behind the obstacles and deep in the shelf. Some scenes are presented in Fig. \ref{fig:instances}.

\begin{figure}[ht]
    \begin{center}
    \begin{overpic}[width=\columnwidth]{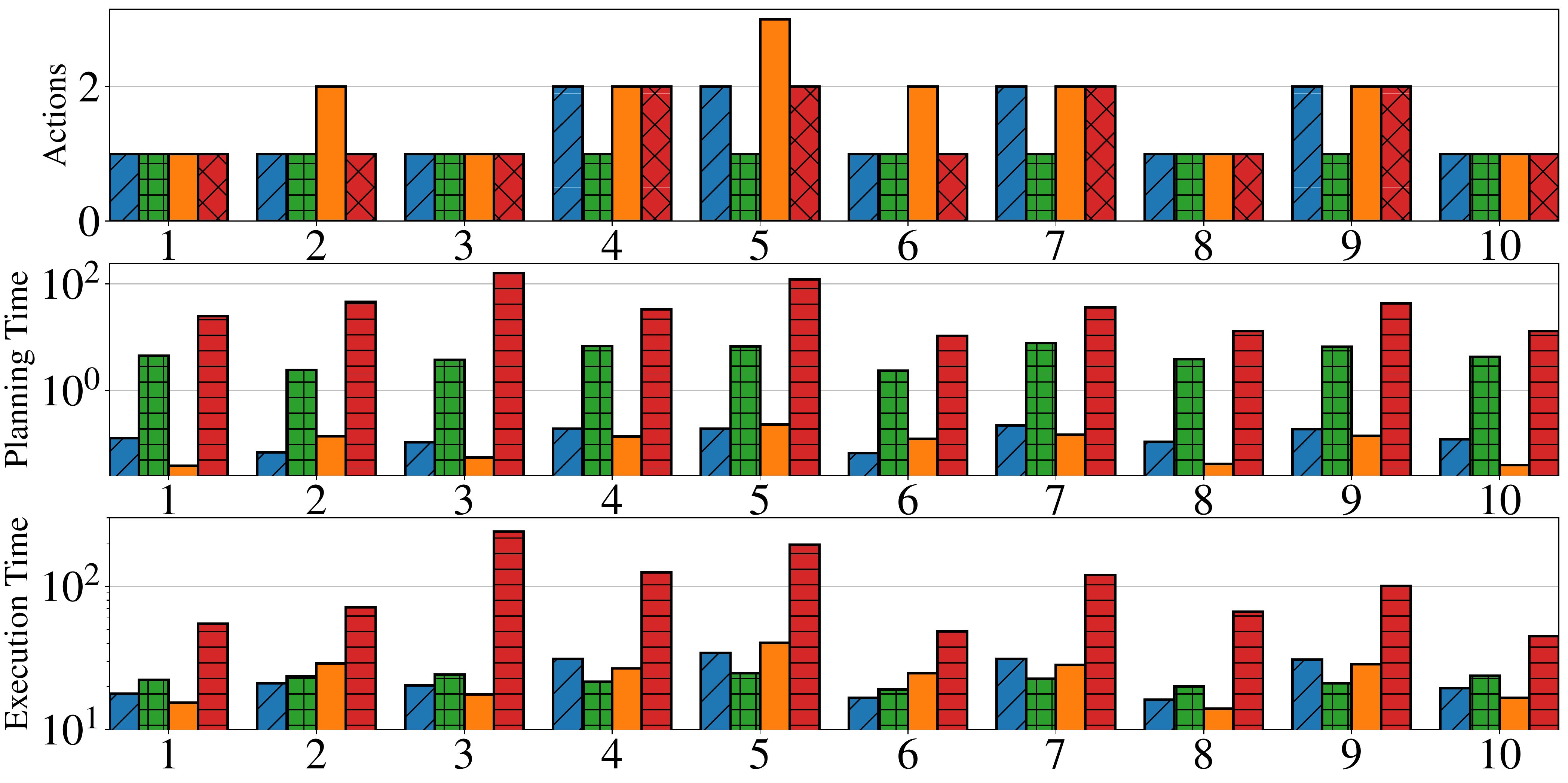}
    \put(24,-16){\includegraphics[width=0.55\columnwidth]{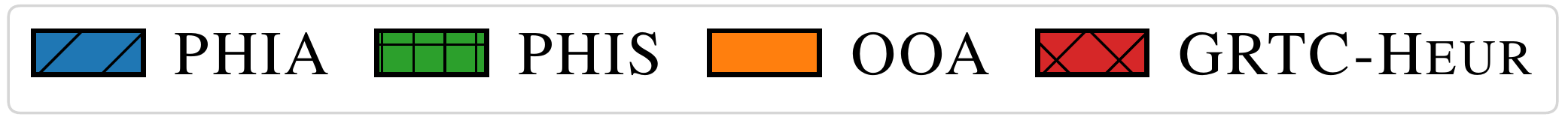}}
    \end{overpic}
    \includegraphics[width=0.6\columnwidth]{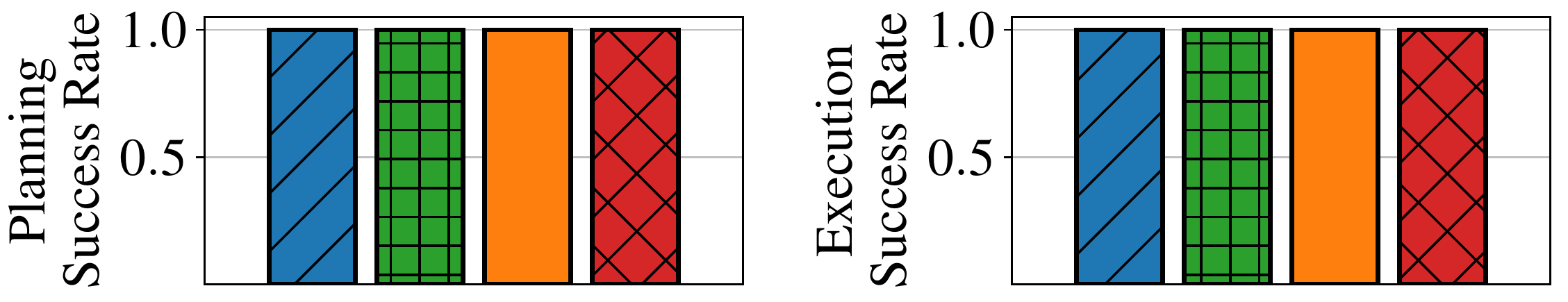}
    \end{center}
    \caption{Ten simulations examples for the simple case where the number of objects is four, see top row of Fig. \ref{fig:instances}.}
    \label{fig:easy}
\end{figure}

Fig. \ref{fig:easy} summarizes the results of the experiments for the scenes with three obstacles (easy case, see Fig. \ref{fig:instances}). We have decided to compare all algorithms with a less cluttered environment in order to increase the success rate of the GRTC-Heuristic. The overall success rate is high as expected. The planning time for GRTC-Heuristic, however, is still high since it samples points to perform the pushing action.

\begin{figure}[ht]
    \begin{center}
    \begin{overpic}[width=\columnwidth]{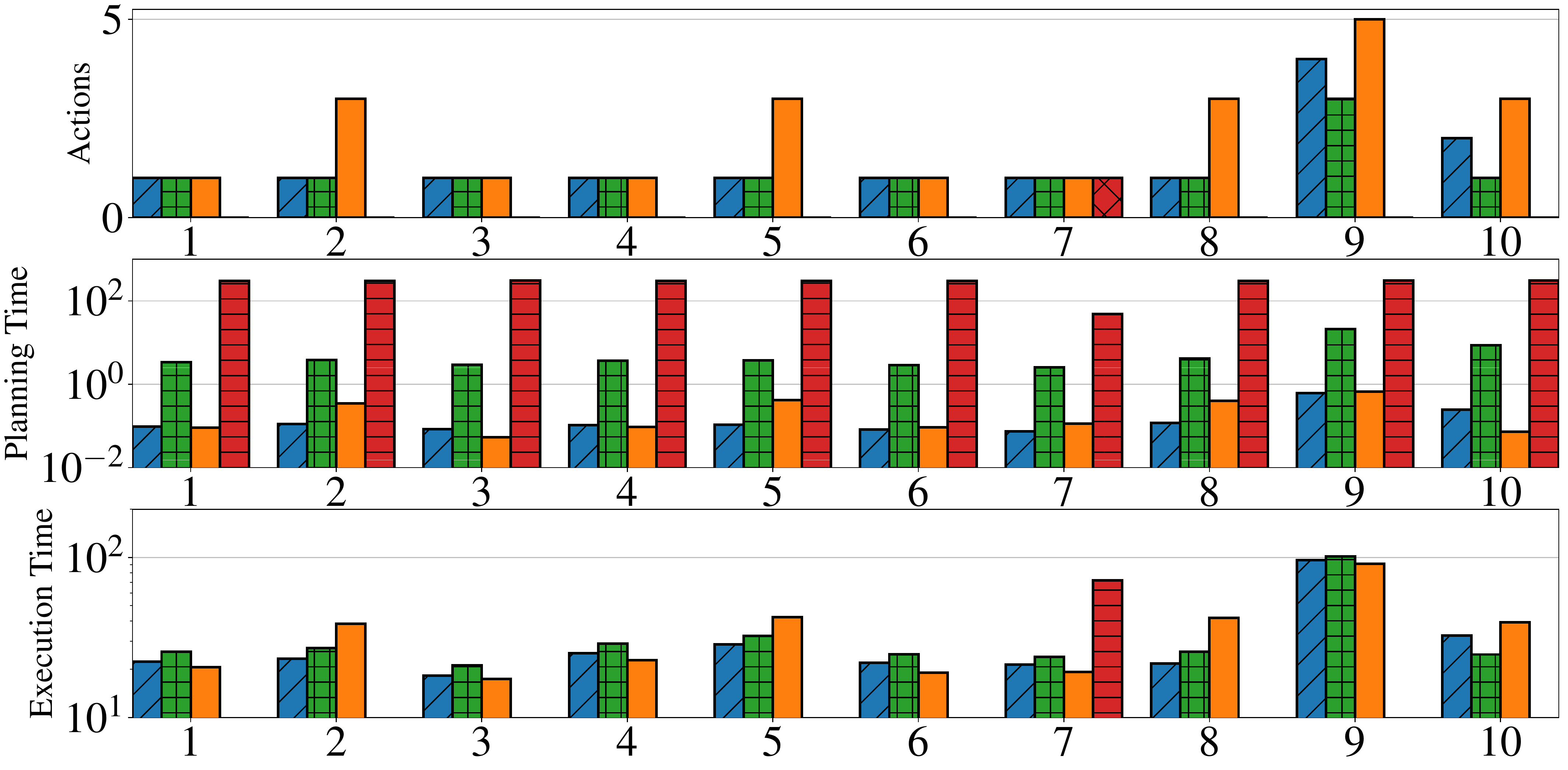}
    \put(24,-16){\includegraphics[width=0.55\columnwidth]{figures/fig_legend.pdf}}
    \end{overpic}
    \includegraphics[width=0.6\columnwidth]{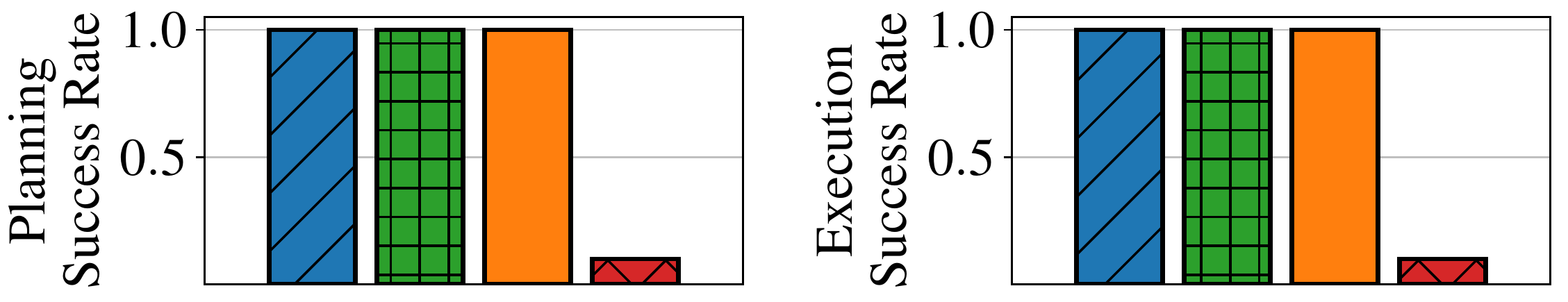}
    \end{center}
    \vspace{.05in}
    \caption{Ten simulation examples from a prior work \cite{papallas2020non}. Execution time is always zero when the planning time reaches 300s.}
    \label{fig:pappa}
\end{figure}

The comparison results with the scenes provided in related work \cite{papallas2020non} is shown in Fig. \ref{fig:pappa}. The success rate for GRTC-Heuristic is low and it agrees with the experiments shown in the related work \cite{papallas2020non}. Note that the number of actions for the algorithm OOA may be high but it eventually achieves success. When the number of actions is equal between PHIA and OOA, it is possible to see that the planning time for PHIA is slightly higher than OOA. The reason behind this difference is the computation time to perform the topological data analysis. Nevertheless, when the number of actions is different between PHIA and OOA, the planning time for OOA increases since for each configuration $\config$ it has to find the closest obstacle to be pushed away from the path region $\pathregion(\config)$.

\begin{figure}[ht]
\vspace{1mm}    \includegraphics[width=\columnwidth]{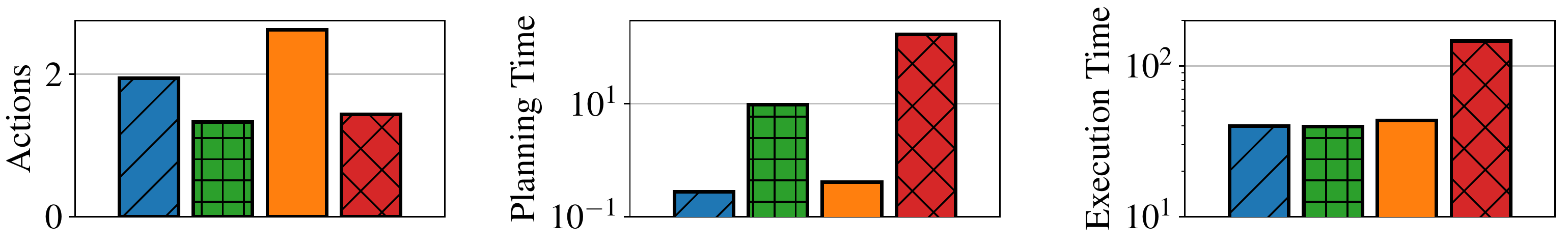}
    \begin{center}
    \begin{overpic}[width=0.6\columnwidth]{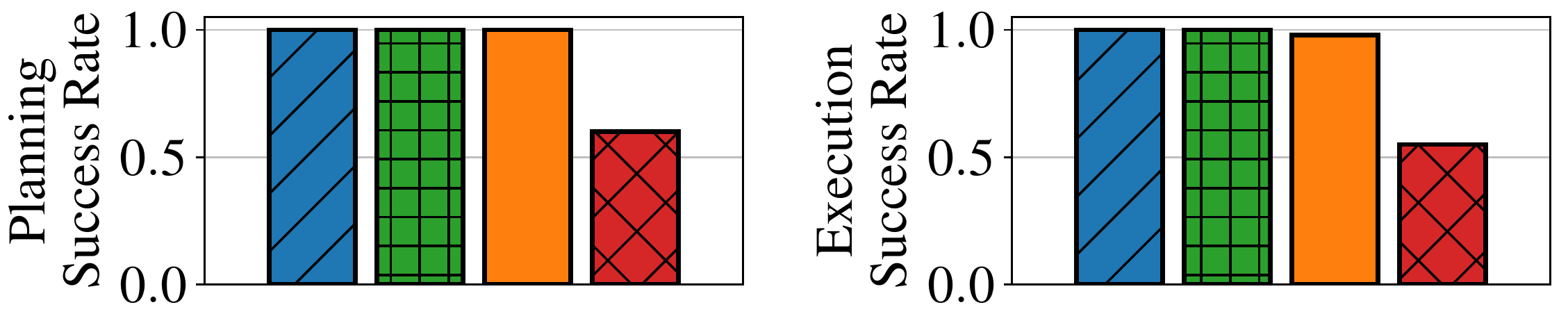}
    \put(6,-7){\includegraphics[width=0.55\columnwidth]{figures/fig_legend.pdf}}
    \end{overpic}
    \end{center}
    \vspace{.05in}
    \caption{One hundred random cases, where each bar represents the average of the 100 random instances for each algorithm.}
    \label{fig:rand}
    \vspace{.1in}
\end{figure}

The results from the 100 random scenes shown in Fig. \ref{fig:rand} further confirm the conclusions presented in previous paragraphs. The average of actions proposed by GRTC-heuristic is shown only for the successful cases. Observe that the average number of actions performed by PHIS is lower than the other methods. This comes, however, with a significant increase in the planning time but not enough to pass the planning time of the GRTC-heuristic. In summary, the conclusion we can draw from the experiments is that PHIA has the best planning time and PHIS is the best in terms of number of actions.

\section{Conclusion and Future Work}

We presented an application of topological data analysis for non-prehensile manipulations in clutter. Our simulated experiments show that the algorithms based on persistent homology are more successful and faster in finding solutions than the baselines and alternatives from the literature. The experiments indicate that the topological data analysis is cheaper to compute. Its use did not notably increase planning time. Instead, it decreased planning time since fewer actions needed to be performed.

The success of the experiments motivates to further explore the topological method for non-prehensile manipulation. For instance, we simplified the problem by considering the objects in a cylindrical shape for experiments; however, the framework can be applied to a more general setup. One direction is to use  persistent homology to also obtain information about the complementary region of the path region in the workspace, this will be more crucial when the density of the workspace gets high. Another direction to consider is to use the Vietoris–Rips complex to obtain the shape of connected components to increase the efficiency of each pushing action. 

For future work, perception should also be taken into account since the connected components that are persistent are robust. Therefore it can provide a good approach for pushing actions when the uncertainty of the pose of the objects is high but bounded. Different types of manipulations may be useful to consider, such as pick and place actions to separate highly dense clusters of objects.

%%%%%%%%%%%%%%%%%%%%%%%%%%%%%%%%%%%%%%%%%%%%%%%%%%%%%%%%%%%%%%%%%%%%%%%%%%%%%%%%

\newpage

\bibliographystyle{format/IEEEtran}
\bibliography{bib/c}

\end{document}